\documentclass[10pt, a4paper]{article}
\usepackage[final]{lrec2026} % this is the new style
\usepackage{iftex}
\ifLuaTeX
    \usepackage[english]{babel} % Set the main language (e.g., English)
    \babelprovide[import=zh-Hans, onchar=ids fonts]{chinese} 
    \babelfont[chinese]{rm} {Noto Serif CJK SC}  
\else
    \usepackage{CJKutf8}
\fi

\usepackage[normalem]{ulem}
\usepackage{comment}
\usepackage{amsmath}
\usepackage{threeparttable}
\usepackage{array}
\usepackage{multirow}

\newcommand{\unrelated}{\emph{Unrelated}}
\usepackage{todonotes}

\title{Multilingual Target-Stance Extraction}

\name{Ethan Mines, Bonnie Dorr} 

\address{University of Florida, Gainesville, USA \\
         \{ethanlmines, bonniejdorr\}@ufl.edu\\}

\abstract{
Social media enables data-driven analysis of public opinion on contested issues.
Target-Stance Extraction (TSE) is the task of identifying the target discussed in a document and the document's stance towards that target.
Many works classify stance towards a \textit{given} target in a multilingual setting, but all prior work in TSE is English-only.
This work introduces the first multilingual TSE benchmark, spanning Catalan, Estonian, French, Italian, Mandarin, and Spanish corpora.
It manages to extend the original TSE pipeline to a multilingual setting without requiring separate models for each language.
Our model pipeline achieves a modest F1 score of 12.78, underscoring the increased difficulty of the multilingual task relative to English-only setups and highlighting target prediction as the primary bottleneck.
We are also the first to demonstrate the sensitivity of TSE's F1 score to different target verbalizations.
Together these serve as a much-needed baseline for resources, algorithms, and evaluation criteria in multilingual TSE.
 \\ \newline  }
 
\begin{document}

\maketitleabstract

\section{Introduction}
Stance detection is the task of predicting an author's opinion or stance towards a particular target.
Potential targets can include political candidates \citelanguageresource{li-etal-2021-p}, COVID-19 policies \citelanguageresource{glandt-etal-2021-stance}, and corporate mergers \citelanguageresource{conforti-etal-2020-will}, among others studied in recent stance detection work.

Most contemporary stance detection systems rely on
%Today's stance detection algorithms are typically 
machine learning models.
One can either train a separate model for each possible target \cite{zarrella-marsh-2016-mitre} or a general model that takes a representation of the target as part of its input \cite{augenstein-etal-2016-stance, xu-etal-2018-cross}.
The latter variant allows one to perform zero-shot stance detection: detecting the stance towards targets not seen during training \citelanguageresource{allaway-mckeown-2020-zero}.

\begin{figure}
\begin{center}
    \includegraphics[width=\columnwidth]{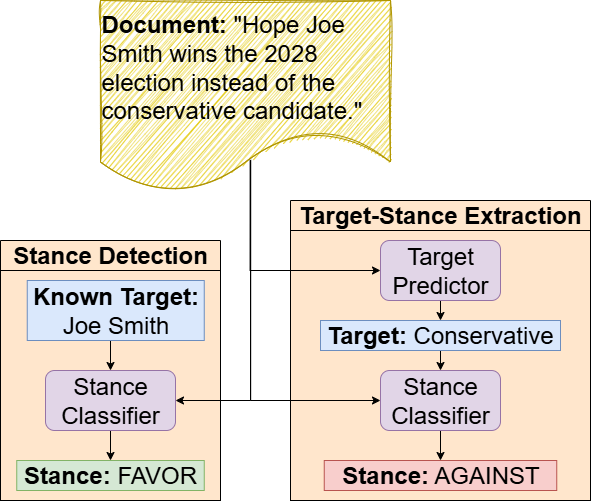}
    \caption{Stance Detection versus Target-Stance Extraction}
    \label{fig:tse_overview}
\end{center}
\end{figure}

In either case, one must already know the target of the author's writing prior to detecting the stance. 
This generally requires either
%This at least requires careful 
keyword-based retrieval of relevant social media posts
%searches of social media for documents relevant to the target(s) of interest 
\citelanguageresource{conforti-etal-2020-will} or explicit human annotation of the targets.
%, if not explicit human annotations of the targets.
\citet{li-etal-2023-new} introduce Target-Stance Extraction (TSE), the task of first predicting the target of the document and then the document's stance toward that target.
As illustrated in Figure~\ref{fig:tse_overview}, TSE is more useful to stakeholders because it eliminates the need to predefine relevant targets.

The primary contributions of this study 
%reported herein 
are (1) generalizing 
%generalization of 
the original TSE algorithm to a multilingual setting and (2) introducing a benchmark dataset to facilitate further progress in multilingual TSE.
While stance detection has been performed for languages other than English, all existing TSE works focus exclusively on English.
A multilingual TSE algorithm would allow decision makers to ascertain public opinion across cultures.
We adapt Target-Stance Extraction to a multilingual scenario, including  Romance languages, Estonian, and Mandarin.
The accompanying multilingual TSE benchmark dataset combines existing stance detection corpora for these languages.
Our generalized algorithm predicts both the target and stance for a sample, without relying on the language itself to determine the prediction.\footnote{Our anonymized source code is available at 
\url{https://osf.io/v65nw/overview?view_only=9bea7b37511c454bb44f844fe97936fd}.}

\section{Related Work}

This section reviews stance detection in general (\S\ref{ssec:rel_stancedet}), multilingual stance detection (\S\ref{ssec:rel_multistancedet}), and the more general TSE task %of Target-Stance Extraction 
(\S\ref{ssec:rel_tse}).

\subsection{Stance Detection} \label{ssec:rel_stancedet}

The first widely adopted stance detection benchmark is the SemEval2016 Task 6 stance dataset, consisting of six targets relevant to U.S. politics \citelanguageresource{mohammad-etal-2016-semeval}.
Early studies train separate classifiers for each target \cite{zarrella-marsh-2016-mitre,wei-etal-2016-pkudblab}. 
%Initial works train separate classifiers for each of the targets .
Later work introduces cross-target stance detection, where a model trained on one target classifies stance toward a related target \cite{xu-etal-2018-cross, liang_target-adaptive_2021}.
%Later studies introduce the notion of cross-target stance detection: using a classifier trained specifically for one target to classify stance towards another, related target 
The next step beyond this is zero-shot stance detection (ZSSD), which evaluates a model on targets unrelated to those seen in training.
%evaluating a stance classifier on targets that have no specific relation to those seen in training.
The VAST dataset \citelanguageresource{allaway-mckeown-2020-zero} is the standard benchmark for ZSSD, though the newer EZ-Stance dataset is a viable alternative \citelanguageresource{zhao-caragea-2024-ez}.
More recent has been the development of multimodal stance corpora \citelanguageresource{niu_multimodal_2024,liang-etal-2024-multi}.

Since the introduction of BERT \cite{devlin2019bert}, it is common practice to fine-tune a pretrained BERT model for stance classification \cite{liang-etal-2022-jointcl,luo-etal-2022-exploiting,zhao-caragea-2025-bilingual}.
The model typically takes a concatenation of the target and the document:
\begin{equation*} \texttt{ [CLS] target [SEP] document [SEP]  } \end{equation*}
The final hidden state of the \texttt{[CLS]} token is passed to a classifier head that assigns the appropriate stance label.

Beyond choice of model, many works introduce adversarial learning objectives to improve the hidden representations for targets and/or documents \cite{wei_modeling_2019,allaway-etal-2021-adversarial,liang-etal-2022-jointcl}.
Others retrieve supplementary background knowledge---typically Wikipedia descriptions of the target---to enrich contextual understanding \cite{zhu_enhancing_2022, he-etal-2022-infusing, li-etal-2023-stance}.
Still others use knowledge graphs like ConceptNet to augment their models with symbolic knowledge \cite{liu-etal-2021-enhancing,luo-etal-2022-exploiting,chen_sentkb-bert_2024}.

As in these related studies, the stance classification stage of our pipeline uses a BERT model that predicts stance given the concatenation of the target and document.
In contrast, we also train a separate model for predicting the target prior to performing stance classification.

\subsection{Multilingual Stance Detection} \label{ssec:rel_multistancedet}

Non-English stance detection datasets vary widely in scope, both regarding language and their targets.
Many resources are limited to a single issue in a single language, such as immigration to Estonia \citelanguageresource{mets_automated_2024}.
Broader collections contain a small set of targets relevant to a specific nation, such as French electoral candidates \citelanguageresource{lai_multilingual_2020}, but are still limited to one language.
Truly multilingual stance corpora---such as X-Stance \citelanguageresource{vamvas_xstance} and Stanceosaurus \citelanguageresource{zheng-etal-2022-stanceosaurus}---span multiple languages and target domains, enabling cross-lingual transfer.
%include multiple languages as well as targets, allowing for cross-lingual transfer.

Regarding algorithms, the creators of X-Stance show that a pretrained multilingual BERT model \cite{devlin2019bert} can be fine-tuned to classify multilingual stance samples \citelanguageresource{vamvas_xstance}.
\citetlanguageresource{lai_multilingual_2020} use a feature-based approach, studying the use of different machine learning features for stance detection in the political domain across five languages.
\citet{hardalov_few-shot_2022} apply sentiment analysis to multilingual Wikipedia articles to create silver training data for their stance classifier.
They create a multilingual stance benchmark by combining many existing stance datasets and demonstrate the feasibility of few-shot training a stance classifier for a new language.
\citet{zhao-caragea-2025-bilingual} similarly create a bilingual benchmark by combining the Mandarin C-Stance \citelanguageresource{zhao-etal-2023-c} and English EZ-Stance \citelanguageresource{zhao-caragea-2024-ez} datasets and evaluating cross-lingual transfer.

Building on these multilingual corpora, we construct a new TSE benchmark that extends prior stance detection efforts.
%We leverage several of the non-English stance corpora to create a multilingual TSE benchmark.
As in recent studies, our approach relies on pretrained language models such as BERT, but our focus is %Distinct from prior work, this study addresses 
the more general task of Target-Stance Extraction, not just stance detection alone.

\subsection{Target-Stance Extraction} \label{ssec:rel_tse}

\citet{li-etal-2023-new} introduce the task of Target-Stance Extraction (TSE), predicting both the target of a document and the document's stance toward it.
They adopt two different approaches to predicting targets: Target Classification (TC) and Target Generation (TG).
TC fine-tunes a pretrained BERT model with a classifier head to predict among a fixed pool of targets.
TG uses a BART sequence transduction model \cite{lewis-etal-2020-bart} to generate a free-form target.
For evaluation purposes, the free-form targets are still mapped to a fixed pool of targets using cosine embedding similarity.
Regardless of how targets are predicted, an additional BERT model is fine-tuned to predict stance given a target and a document.
They form their benchmark dataset by collating four existing English stance datasets.

\citet{yan_bcltc_2025} simplify the TC approach by using the same underlying BERT encoder for both target prediction and stance prediction.
They train two different classifier heads for the model: one for predicting the target when given only a document and another for predicting the stance given both a document and a target.

\citet{akash-etal-2025-large} introduce Open-Target Stance Detection (OTSD), a scenario similar to Target-Stance Extraction with Target Generation.
They validate the effectiveness of using large-language models for target prediction and stance classification without any pretraining.
OTSD differs from TSE:
%in its evaluation protocol: 
OTSD emphasizes open-target discovery without a fixed mapping step, relying on human evaluation to verify target quality.
TSE on the other hand requires mapping the predicted target to a pool of predetermined targets during evaluation.
Our study follows the TSE setting.

\begin{figure}[ht]
\begin{center}
    \includegraphics[width=\columnwidth]{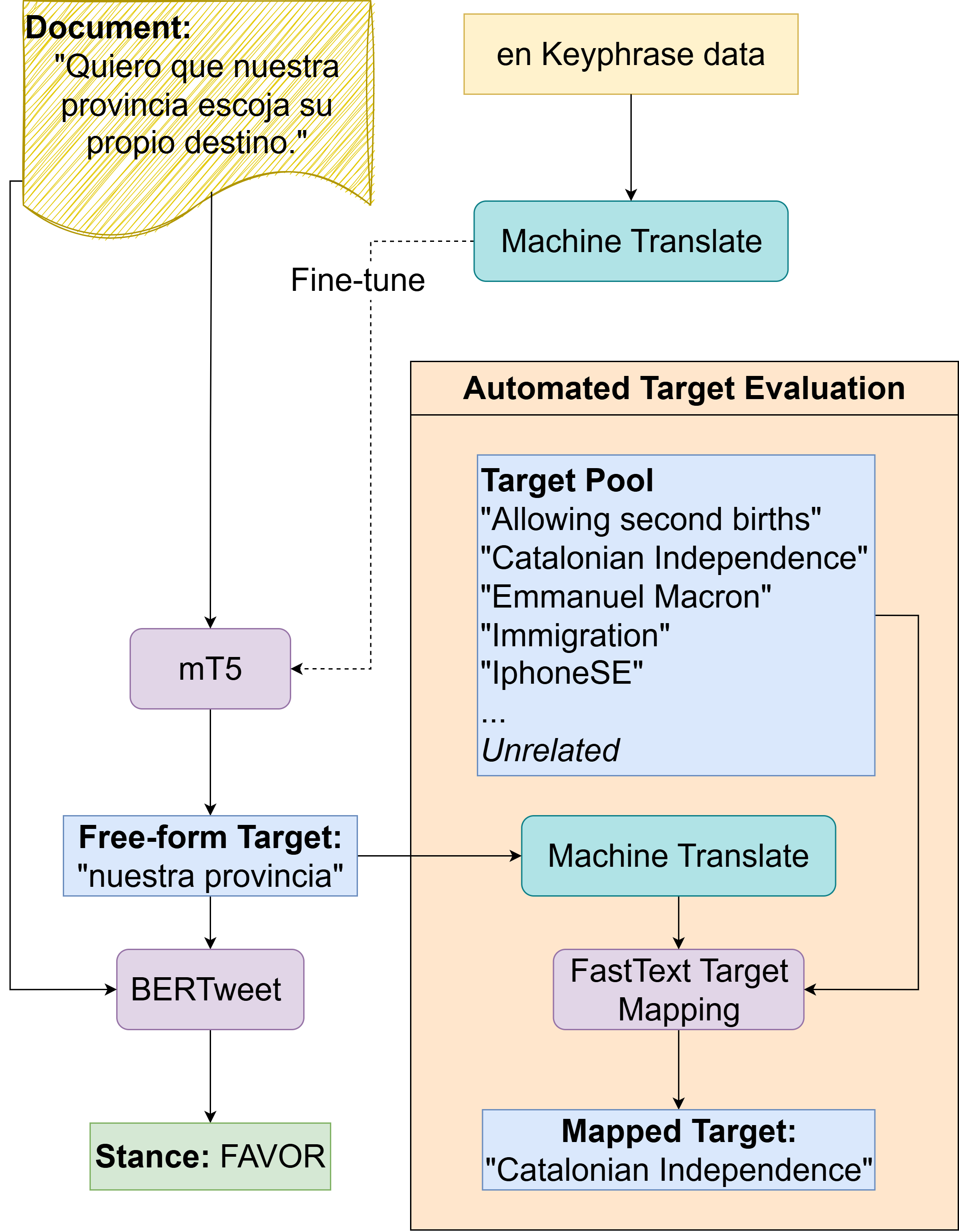}
    \caption{Overview of Multilingual Target-Stance Extraction. A machine-translated keyword corpus is used to fine-tune an mT5 sequence model for target prediction. A BERTweet model predicts the stance given the original document and the target prediction. Using FastText embedding similarity, the free-form predicted target is mapped to a set of fixed targets for evaluation purposes. }
    \label{fig:tse_dataflow}
\end{center}
\end{figure}

\citet{Mather_Dorr_Rambow_Strzalkowski_2021} introduce a novel approach to stance detection that requires no machine learning models but some semi-automated resource construction.
They provide a more explainable grounding for stance as %in the form of 
a belief regarding the target plus a sentiment towards that belief.
While this work does not use such a notion of stance, both works are able to predict targets in addition to stance.

Like \citet{li-etal-2023-new}, we collate existing stance corpora to form a benchmark TSE dataset. 
Distinct from prior multilingual stance work, our study addresses TSE, which requires predicting both targets and stances.
Extending earlier TSE studies, the pipeline operates in a multilingual setting.
Additionally, the Target Generation mode is adopted exclusively for predicting targets.
The non-English stance corpora used here each have very distinct target sets.
A simple classification head (i.e., the task-specific output layer) can easily memorize the mapping between each corpus and its targets---even if all language cues are removed by translating all texts into the same language.

Building on prior work in multilingual stance detection and English-only TSE, the following section describes our design for a multilingual TSE benchmark and pipeline.

\section{Methodology}

Described here are the problem definition for Target-Stance Extraction (\S\ref{ssec:prob_def}), the multilingual TSE algorithm (\S\ref{ssec:tse_pipe}), and the data used for training and evaluation (\S\ref{ssec:data}).

\subsection{Problem Definition} \label{ssec:prob_def}
Consider an input document $x$, a groundtruth (GT) target $t$, and a groundtruth stance $y \in \{\text{Against}, \text{Favor}, \text{Neutral}\}$.
The goal of stance detection is to predict a stance $\hat{y}$ given $(x,t)$.
The goal of target-stance extraction is to predict both $\hat{y}$ and a target $\hat{t}$ given only $x$.

\subsection{TSE Pipeline} \label{ssec:tse_pipe}

We operationalize multilingual TSE with a two-stage pipeline (Figure \ref{fig:tse_dataflow}). 
An mT5 sequence transduction model \cite{xue-etal-2021-mt5} first generates a free-form target for the input document, and a BERTweet \cite{nguyen-etal-2020-bertweet} stance classifier then predicts stance given the document and the generated target.
The mT5 model is fine-tuned on a machine-translated keyphrase-generation corpus, and the stance model is fine-tuned on labeled stance examples.
Due to the keyphrase generation data used to fine-tune the mT5 model, a single output sequence can have multiple keyphrases.
During postprocessing, exact duplicates are removed to ensure unique candidate targets.
%among the keyphrases are removed in postprocessing.

To perform an automated evaluation of generated targets, there needs to be a way to map them to a set of known targets.
Following \citet{li-etal-2023-new}, all possible targets from the stance corpora are collated into a target pool.
To simplify the mapping process, only English verbalizations of the targets are used in this pool.
This choice standardizes embeddings across languages and minimizes translation noise, ensuring that similarity comparisons occur in a single, consistent vector space. 

\begin{table}[t]
\footnotesize
\centering

\begin{threeparttable}
    \begin{tabularx}{\linewidth}{|p{1.615cm}|X|}
        \hline
        Target           & Verbalization \\
        \hline
       catalonia    & Catalonian Independence                                                 \\
       \hline
       immigration  & Immigration                                                             \\
       \hline
       lepen        & Marine LePen                                                            \\
       \hline
       macron       & Emmanuel Macron                                                         \\
       \hline
       sardinia     & Sardinian Independence                                                  \\
       \hline
       firecracker  & Setting off firecrackers during the Spring Festival \tnote{a}           \\
       \hline
       iphone       & IphoneSE \tnote{b}                                                      \\
       \hline
       russia       & Russia's counter-terrorism operations in Syria \tnote{c}                \\
       \hline
       secondbirth  & Allowing second births                                                  \\
       \hline
       shenzen      & Shenzhen bans motorcyles and imposes electricity restrictions \tnote{d} \\
    \hline
    \end{tabularx}
    
    \begin{tablenotes}
        \scriptsize
        \item[a] LLM: ``Firecracker Spring Festival''; Manual: ``Firecrackers''
        \item[b] LLM: ``iPhone SE'' 
        \item[c] LLM: ``Russian counterterrorism in Syria''; Manual: ``Russia'' 
        \item[d] LLM: ``Shenzhen motorcycle electricity''; Manual: ``Shenzhen Laws''
    \end{tablenotes}
    \caption{
        Target labels and their verbalizations.
        Same verbalization is used across all three pools (Full, LLM, Manual) unless indicated otherwise.
    }
    \label{tab:target_ke}
\end{threeparttable}
\end{table}

The free-form target is translated by an MT model into English.
A set of pretrained FastText embeddings \cite{bojanowski-etal-2017-enriching} are used to map this translated target to a pool entry.
For each target in the pool, the cosine similarity is computed between the prediction's embedding and the target's embedding.
If no cosine similarity exceeds a predefined threshold $\tau$, the prediction is mapped to a special \unrelated\ target.  
Otherwise, the prediction is mapped to the target with the highest cosine similarity.
In practice, the sequence transduction model returns multiple predictions in its output sequence;
the single prediction with the highest similarity to any pool entry is retained. 

Note that while the mapped target is the target used when evaluating model performance, an end user of the system would only see the generated free-form target.
Additionally, it is the free-form target that is provided to the BERTweet model for stance classification.

\subsection{Data} \label{ssec:data}

\begin{table}[t]
\small
\begin{center}
    \begin{tabular}{|l|l|r|r|r|}
\hline
Lang & Target & Against & Favor & Neutral \\
\hline
ca & catalonia & 3988 & 3902 & 2158 \\
    & \unrelated & -- & -- & 2087 \\
\hline
es & catalonia & 4105 & 4104 & 1868 \\
    & \unrelated & -- & -- & 2093 \\
\hline
et & immigration & 1175 & 489 & 1597 \\
        & \unrelated & -- & -- & 677 \\
\hline
fr & macron & 308 & 91 & 131 \\
    & lepen & 466 & 65 & 55 \\
     & \unrelated & -- & -- & 231 \\
\hline
it & sardinia & 1770 & 785 & 687 \\
    & \unrelated & -- & -- & 673 \\
\hline
zh
    & firecracker & 250 & 250 & 100 \\
    & iphone & 209 & 245 & 146 \\
    & russia & 250 & 250 & 100 \\
    & secondbirth & 200 & 260 & 140 \\
    & shenzhen & 300 & 160 & 126 \\
    & \unrelated & -- & -- & 620 \\
\hline
    \end{tabular}
    \caption{Sample counts by language, target label, and stance label (Against, Favor, Neutral). \unrelated\ samples may only be Neutral.}
    \label{tab:corp_stats}
\end{center}
\end{table}

We construct a multilingual TSE benchmark consisting of Spanish, Catalan, French, Italian, Estonian, and Mandarin samples.
Samples labeled Favor, Against, or Neutral are drawn from the following existing stance corpora: 
\begin{tabular}{p{3.2in}}
 $\bullet$ Catalonian Independence Corpus (\textbf{cic}) \newline
 \mbox{~~~} \textit{\citelanguageresource{zotova-etal-2020-multilingual}}\\
 $\bullet$ French Election Corpus (\textbf{e-fra})  \newline
 \mbox{~~~} \textit{\citelanguageresource{lai_multilingual_2020}}\\
 $\bullet$ SardiStance Corpus (\textbf{sardi}) \newline 
 \mbox{~~~} \textit{\citelanguageresource{basile_sardistance_2020}}\\
 $\bullet$ Estonian Immigration Corpus (\textbf{et-imm}) \newline 
 \mbox{~~~} \textit{\citelanguageresource{mets_automated_2024}}\\
 $\bullet$ NLPCC 2016 Stance Corpus (\textbf{nlpcc}) \newline
 \mbox{~~~} \textit{\citelanguageresource{xu_overview_2016}}
\end{tabular}

For the mapping process, the available targets from each stance corpus are combined to form a pool.
Table~\ref{tab:target_ke} lists each target label along with its corresponding textual verbalization.
Since the verbalizations used for targets can greatly influence the outcome of the mapping, three target pools are considered: Full, LLM, and Manual.
Full refers to the original, precise target names from each dataset.
LLM verbalizations are obtained by prompting the 20b-parameter OSS ChatGPT \cite{openai2025gptoss120bgptoss20bmodel} to trim each target name to three words or fewer.
Manual verbalizations are created by manually trimming the target names to shorter phrases. For most targets, the verbalization is identical across all pools.

An additional \unrelated\ class is used to provide true negatives for TSE evaluation and to avoid forcing a spurious target match when a document does not express stance toward any target. 
\unrelated\ samples are derived as follows: for Estonian, additional non-target posts from \textbf{et-imm} are used; 
for Mandarin posts are drawn from C-Stance \citelanguageresource{zhao-etal-2023-c}, with target annotations ignored; and 
for Catalan, French, Italian, and Spanish, sentences are sampled from GlobalVoices \citelanguageresource{tiedemann-2012-parallel}.
For each language, the \unrelated\ portion is set to approximately 17\% of the samples to align with prior TSE practice \citep{li-etal-2023-new}.
Table~\ref{tab:corp_stats} provides the number of samples for each language, target, and stance label.

Beyond the TSE benchmark data, 64,000 English samples are drawn from GlobalVoices for training the FastText embeddings.
This size is comparable to the $\sim$50,000 samples used by \citet{li-etal-2023-new} to train their embeddings.
The keyphrase generation corpus KPTimes \cite{gallina-etal-2019-kptimes} is used for target-predictor training.
All the development samples and 1,000,000 training samples are translated into the six languages (1/6 of the corpus for each language).

\section{Experiments}

Evaluation covers target prediction, stance detection, and overall performance. 
For target prediction, we report per-class F1 
%score is reported, 
along with the micro- and macro-averages 
%average 
($F_{\text{mic}}$,
%and the macro average 
$F_{\text{mac}}$) across classes.

Following \citet{li-etal-2023-new}, the macro average of F1 score over the Favor and Against classes, as shown in  Eq.~\ref{eq:favg}, is used to evaluate the stance classifier.
This metric implicitly captures performance on the Neutral class.

\begin{equation}  
    F_{\text{avg}} = \frac{F_{\text{against}} + F_{\text{favor}}}{2}
    \label{eq:favg}
\end{equation}

Any samples with the \unrelated\ target are excluded from this evaluation as their stance is irrelevant.
Note there are still samples that have a Neutral label but not the \unrelated\ target.
$F_{\text{avg}}$ is reported for each (language, target) pair, as well as a macro average $F_{\text{mac stance}}$.

To evaluate the TSE system as a whole, the TSE F1 metric from \citet{li-etal-2023-new} is used.
Any sample with the \unrelated\ target is defined to be a negative sample; otherwise it has an actual target and is considered a positive sample.
The system must predict the correct target and stance for a positive sample to obtain a true positive.
Predicting any target but \unrelated\ for a negative sample yields a false positive.
Predicting the \unrelated\ target for a positive sample yields a false negative.
Predicting the wrong target or wrong stance for a positive 
sample counts both as a false positive and a false negative.
From these definitions precision, recall, and F1 score can be computed.
We report global, per-language, and macro-averaged F1 scores %A global F1 score, an F1 score for each language subset, and a macro average over the language scores 
($F_{\text{mac tse}}$). % are reported.

FastText embeddings (256-d) are trained for 500 epochs on the English GlobalVoices data.
For target generation, a pretrained mT5 model \cite{xue-etal-2021-mt5} from HuggingFace\footnote{\url{https://huggingface.co/google/mt5-base}}
is fine-tuned on the machine-translated keyphrase generation corpus for 24 hours with a batch size of 32.
Validation occurs every 500 batches, and the checkpoint with the lowest validation cross-entropy is retained.
Based on preliminary experiments, we choose a cosine similarity threshold of $\tau = 0.35$ for mapping free-form targets to fixed ones.

For stance classification, a BERTweet model \cite{nguyen-etal-2020-bertweet} from HuggingFace\footnote{\url{https://huggingface.co/vinai/bertweet-base}} is fine-tuned
for five epochs with a batch size of 32; training takes approximately 20 minutes.
The checkpoint with the highest validation $F_{\text{avg}}$ is retained and---following \citet{li-etal-2023-new}---\unrelated\ samples are excluded from stance training.

\begin{table}
\begin{center}
    \begin{tabular}{|l|r|r|r|r}
        \hline
        Target    & Full & LLM & Manual \\
        \hline
        catalonia & 17.96 & 17.97 & \textbf{17.99} \\
        immigration & \textbf{42.60} & 25.66 & 25.31 \\
        lepen     & 34.77 & 34.73 & \textbf{34.81} \\
        macron & 17.96 & 17.97 & \textbf{17.99} \\
        sardinia & 19.86 & \textbf{20.78} & 20.72 \\
        firecracker & \textbf{4.25} & 2.47 & 2.87 \\
        iphone & \textbf{42.60} & 25.66 & 25.31 \\
        russia & 6.70 & 30.37 & \textbf{30.43} \\
        shenzhen & 18.58 & \textbf{38.62} & 0.00 \\
        secondbirth & 0.58 & 0.56 & \textbf{0.59} \\
        \unrelated & 22.19 & \textbf{27.81} & 25.37 \\
        \hline
        $F_{\text{mic}}$ & 20.94 & \textbf{24.19} & 22.75 \\
        $F_{\text{mac}}$ & 22.14 & \textbf{27.82} & 21.24 \\        
       \hline
    \end{tabular}
    \caption{Target prediction F1 score using different target pools. Results averaged across five folds. }
    \label{tab:target_results}    
\end{center}
\end{table}

Five-fold cross-validation is performed, and results are averaged across folds.
Splits are stratified to preserve---within each fold---the proportions of language, target, and stance labels observed in the full dataset.
%Note that c
Cross-validation applies only to the stance data (not the keyphrase data), so the same keyphrase 
%generation
model is used across all five folds.
All machine translation, both of the keyphrase corpus and of the generated targets, is done with a pretrained M2M100 model\footnote{\url{https://huggingface.co/facebook/m2m100\_418M}} \cite{fan-m2m100}.

One NVIDIA B200 GPU is used for keyphrase model fine-tuning, target prediction, and target translation; five NVIDIA L4 GPUs (one for each fold) are used for the remainder of the pipeline.

\section{Results and Analysis}

\begin{table*}[h]
\begin{center}
\begin{tabular}{|l|c|c|c|c|c|c|}
    \hline
    &  \multicolumn{2}{|c|}{Full} & \multicolumn{2}{|c|}{LLM} & \multicolumn{2}{|c|}{Manual} \\
                 \hline
                 &  Mapped     &  GT   &    Mapped &    GT      &  Mapped  & GT     \\
    \hline        
    ca           &   8.25    & \underline{74.45} &   \textbf{8.72}  &  74.04     &  8.69  & 74.33  \\
    es           &   8.24    & \underline{72.25} &   8.43  &  71.79     &  \textbf{8.52}  & 71.96  \\
    et           &  36.25    & \underline{58.54} &  40.98  &  57.74     & \textbf{41.31}  & 57.47  \\
    fr           &  19.56    & 70.43 &  19.60  &  67.72     & \textbf{20.54}  & \underline{70.49}  \\
    it           &   8.63    & 56.60 &   \textbf{9.76}  &  56.14     &  9.60  & \underline{57.00}  \\
    zh           &  16.80    & 45.75 &  \textbf{18.14}  &  46.02     & 13.36  & \underline{46.61}  \\
    \hline
    All          &  12.78    & 67.22 &  \textbf{13.85}  &  \underline{66.73}     & 13.43  & 67.11  \\
    $F_{\text{mac tse}}$ &  16.29    & \underline{63.00} &  \textbf{17.60}  &  62.24     & 17.00  & 62.98  \\
    \hline
\end{tabular}
\end{center}    
\caption{TSE F1 scores for predictions by target pool (Full, LLM, Manual), comparing mapped and GT targets. Results averaged across five folds. The best result for mapped targets is \textbf{bold} while that for GT targets is \underline{underlined}.}
\label{tab:tse_results}
\end{table*}

The target prediction accuracy for individual partitions of the dataset is shown in Table \ref{tab:target_results}.
Across target pools, performance is lower for the targets \emph{firecracker} and \emph{secondbirth}.
%Using a shorter
Shorter verbalizations improve performance
%does substantially improve performance 
for the \emph{russia} target, while results for the \emph{shenzen} legislation target vary 
%more widely. 
considerably across pools.
Performance varies
%Note the performance can vary 
even for targets sharing identical 
%with the same 
verbalizations across pools (e.g., \emph{immigration}).
%, like \emph{immigration}.

Language consistency between input documents and generated targets is further verified using the Lingua language detection library \cite{staab_lingua_2025}.
Table \ref{tab:lang_acc} shows the per-language match rate and its macro average $\text{Avg}_{\text{lang}}$. 
For completeness, every target candidate in the model's output sequence is evaluated, not just the one with that scores the highest embedding similarity with a fixed target.
This evaluation occurs before target mapping, so target pools are not yet applied.
Mandarin scores highest because its distinct script provides a clear signal to the language detector, while Catalan scores lowest primarily due to the Lingua model classifying the output as one of the other Romance languages.
Overall accuracy supports training the keyphrase generator on machine-translated data.

Table \ref{tab:stance_results} reports stance classifier performance across different (language, target) pairs.
Results are more consistent than for target prediction %because 
since the classifier trains directly on stance labels,
%is trained directly on stance labels, 
whereas the target predictor learns from
%is trained on 
a keyphrase-generation corpus, not
%rather than 
GT targets.

Table \ref{tab:tse_results} shows Target-Stance Extraction performance by language.
Under the TSE F1 scheme, a sample counts as a 
%the system obtains a true positive by correctly 
true positive only when 
%predicting 
both the target and stance are correctly predicted for a non-\unrelated\ case.
%sample.
Following \citet{li-etal-2023-new}, a ceiling condition
%is reported in which GT targets 
replaces predicted targets with GT targets.
Performance is lowest for Catalan, Spanish, and Italian, reflecting poor recall on
%due to the system having poor recall for the targets 
\emph{catalonia} and \emph{sardinia}.
The baseline achieves F1 scores between $0.10$ and $0.20$, whereas models on the original English TSE benchmark report F1 scores between $0.30$ and $0.40$ \cite{li-etal-2023-new}.
This gap illustrates the challenge of
%difficulty of 
multilingual TSE and identifies target prediction as the bottleneck.

\begin{table}
\small % Needed this to stay in the hbox

\begin{center}
\begin{tabularx}{\columnwidth}{
    |>{\hsize=0.97\hsize\linewidth=\hsize}X
    |>{\hsize=0.97\hsize\linewidth=\hsize}X
    |>{\hsize=0.97\hsize\linewidth=\hsize}X
    |>{\hsize=0.97\hsize\linewidth=\hsize}X
    |>{\hsize=0.97\hsize\linewidth=\hsize}X
    |>{\hsize=0.97\hsize\linewidth=\hsize}X
    |>{\hsize=1.18\hsize\linewidth=\hsize}X|
}
    \hline
    ca & es & et & fr & it & zh & $\text{Avg}_{\text{lang}}$ \\
    \hline
    71.28 &
    78.59 &
    87.14 &
    77.76 &
    83.14 &
    91.90 &
    81.64 \\
    \hline
\end{tabularx}
\end{center}
\caption{Language match rate of generated targets, averaged across five folds.}
\label{tab:lang_acc}
\end{table}

\begin{table}[t]
\begin{center}
    \begin{tabular}{|l|r|r|r|r}
        \hline
        Lang-Target    & Full  & LLM    & Manual \\
        \hline
        ca-catalonia   &  75.50 & 75.08 & \textbf{75.73} \\
        es-catalonia   &  \textbf{69.85} & 69.08 & 69.36 \\
        et-immigration &  27.25 & \textbf{28.64} & 26.83 \\
        fr-lepen       &  49.33 & 51.09 & \textbf{53.29} \\
        fr-macron      &  43.17 & \textbf{45.00} & 40.82 \\
        it-sardinia    &  50.98 & 50.81 & \textbf{51.97} \\
        zh-firecracker &  40.25 & 41.17 & \textbf{41.98} \\
        zh-iphone      &  \textbf{42.01} & 40.76 & 34.53 \\
        zh-russia      &  40.55 & \textbf{44.98} & 40.62 \\
        zh-shenzhen    &  49.92 & \textbf{50.20} & 48.21 \\
        zh-secondbirth &  34.97 & 32.13 & \textbf{41.80} \\
        \hline
        $F_{\text{mac stance}}$      &  47.62 & \textbf{48.09} & 47.74 \\
        \hline
    \end{tabular}
    \caption{$F_{\text{avg}}$ scores for stance models %using different 
    across target pools (five-fold average). 
    %Results averaged across five folds.
    }
    \label{tab:stance_results}    
\end{center}
\end{table}

\begin{table*}[ht]
\scriptsize{
\begin{tabular}{p{.09in}p{2.06in}>{\raggedright\arraybackslash}p{1.045in}>{\raggedright\arraybackslash}p{1.09in}>{\raggedright\arraybackslash}p{1.08in}}
        \hline
\hspace*{-.03in}
        \textbf{Lang} & \textbf{Document} & \textbf{Generated Targets} & \textbf{Mapped Target} & \textbf{Groundtruth Target} \\
        \hline
        ca & \includegraphics[height=0.85em]{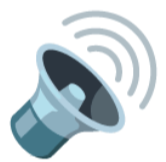}\rule{0pt}{2.0ex}Trapote insisteix en que era el cap de secció qui decidia en quines localitats s'intervenia \#JudiciProcés     & \uline{super bowl};trapote; futbol;nfl;super bowl 2015 & \unrelated & Catalonian Independence \\
        \hline
        es & 4. Y si Arrimadas no quiere ir de dos y aspira al 1. Un discurso de la primera mujer presidenta de España. La ambición de Rivera es desmesurada, aunque quizás fuera el camino para tocar poder. & \uline{política y gobierno};rivera mariana;elecciones presidenciales de 2008 & Russia's counter-terrorism operations in Syria & Catalonian Independence \\
        \hline
        et & Tuneesia ja Liibüa on peamised punktid, kust migrandid asutavad end ohtlikule mereretkele Põhja-Aafrikast Euroopasse, eelkõige Itaaliasse. & \uline{tuneesia, liibüa, euroopa, rahvusvahelised suhted, rahvusvahelised suhted, lähis-ida ja aafrika rände kriis, euroopa rände kriis, euroopa rände kriis} & Immigration & Immigration \\
        \hline
        fr & screenshot exclusif provenant du téléphone de Marine Le Pen  & \uline{marine le pen};marine le pen communications inc;computer et internet & Marine LePen & Marine LePen \\
        \hline
        it & Datemi un nome! Un solo nome di un leghista che abbia fatto qualcosa di buono o intelligente riconosciuto dal mondo o dagli italiani. Non ce ne. E quindi loro sono semplicemente uno scarto. Una razza stupida. Gente che vale meno dello sterco del mulo di mio nonno. Fanculo & \uline{scienza e tecnologia}; razza e etnia;nomi geografici & \unrelated & Sardinian Independence \\
        \hline
\ifLuaTeX
        zh & 中俄战略伙伴关系，不是少数别有用心的中国人，以及冒充中国人的美日等西方网络水军、第五纵队，所能离间分化的！ & \uline{中俄战略伙伴关系};中国;军事 & Russia's counter-terrorism operations in Syria & Russia's counter-terrorism operations in Syria \\
\else
        zh & \begin{CJK*}{UTF8}{gbsn}中俄战略伙伴关系，不是少数别有用心的中国人，以及冒充中国人的美日等西方网络水军、第五纵队，所能离间分化的\end{CJK*} & \begin{CJK*}{UTF8}{gbsn}\uline{中俄战略伙伴关系};中国;军事\end{CJK*} & Russia's counter-terrorism operations in Syria & Russia's counter-terrorism operations in Syria \\
\fi
        \hline
\end{tabular}
}
\caption{Randomly selected target predictions using the Full pool for verbalizations. The chosen candidate is \uline{underlined}.}
\label{tab:sample_preds}
\end{table*}

Sample target predictions for the system are shown in Table \ref{tab:sample_preds}.
One sample from each language corpus is randomly selected.
None of the generated targets for the Catalan nor the Spanish sample explicitly mention Catalonia, making mapping to \emph{Catalonian Independence} difficult.
The same holds for the Italian sample and \emph{Sardinian Independence}.
The single generated target for the Estonian sample explicitly mentions north African nations as well as a migration crisis, easily mapping to \emph{Immigration}.
For the French sample, the model notes the named entity \emph{Marine LePen} and explicitly generates a corresponding target that is mapped correctly.
One of the generated Mandarin candidates explicitly mentions Russia, allowing it to map to the Russian counterterrorism target.

\section{Conclusions and Future Directions}

This work introduces the first multilingual TSE benchmark and a baseline system spanning six languages.
The baseline serves as a reference point for future TSE models.
Although our approach achieves lower metric values than English-only results \cite{li-etal-2023-new}, the resource establishes a crucial starting point for multilingual TSE.
We are also the first to demonstrate the sensitivity of TSE's main metric to target verbalizations.
Most errors arise from incorrect target predictions; in contrast, stance detection on these corpora is well studied. 

Three complementary avenues for improvement are identified: (a) \textit{Multilingual keyphrase corpora}; (b) \textit{Target mapping and embeddings}; and (c) \textit{Broader evaluation datasets}.  

\textbf{(a) Multilingual keyphrase corpora.}
Fine-tuning the keyphrase model on genuine multilingual keyphrase data---rather than machine-translation text---will likely improve performance.
Available resources exist, though they tend to be domain-specific (e.g., law \citelanguageresource{salaun-etal-2024-europa}; scientific research \citelanguageresource{piedboeuf2022a}). 
As a complementary avenue, the generative target predictor could be replaced with an extractive span model; however, span extraction cannot produce targets that do not appear in the input.

\textbf{(b) Target mapping and embeddings.}
There are other means of strengthening the mapping step, for example, by adopting higher-quality multilingual contextual embeddings for similarity computation \cite{akash-etal-2025-large}. 
When translation remains in the loop, using higher-quality MT can reduce drift. Another option is to skip translation and compare representations directly in a shared multilingual space. 
In parallel, future work could
%there is room to 
systematically study target verbalizations and prompting strategies (including model size) to stabilize the mapping and align generated targets with entries in the pool via contrastive objectives.
%consider contrastive objectives that align generated targets with entries in the target pool.

\textbf{(c) Broader evaluation datasets.}
Adapting zero-shot stance resources to TSE \citelanguageresource{allaway-mckeown-2020-zero,zhao-caragea-2024-ez} would provide broader coverage and stronger generalization checks.
To move beyond a small, fixed target pool, evaluation could %consider
test growing pools, %application of 
metrics at multiple pool sizes, and open-world settings %in which unseen targets are permitted.
that allow unseen targets.

Taken together, these directions provide a concrete path toward stronger multilingual TSE systems and more informative evaluations.

\section*{Limitations}

While our benchmark does cover six different languages, four of them (Catalan, French, Italian, Spanish) are Romance languages, potentially limiting the applicability of our results to other groups of languages.
Even for the same languages, different targets than the ten used in our study could lead to different levels of performance, particularly in the target mapping stage.
Annotating a new multilingual dataset specifically designed for TSE, with a wider variety of languages and targets, would address these issues inherent in collating existing stance corpora.

\section*{Ethical Considerations}
While our benchmark is formed from publicly available stance corpora, we do not redistribute the data and thus adhere to copyright restrictions and social network privacy policies.

Regarding application, target-stance extraction is an excellent tool for good actors investigating public opinion on relevant issues.
Unfortunately, this same tool could enable bad actors on social media to find individuals expressing a particular opinion and target them for harassment, or worse.
One future line of work to mitigate this is adversarial Target-Stance Extraction: altering a document to preserve its meaning while causing a misclassification from the target predictor or stance classifier.

\clearpage % The header was awkwardly on the bottom of the page
\section*{Bibliographical References}\label{sec:reference}

% Some sort of glitch where an extra ``References'' header gets made, even though the template is supposed to suppress that
% Got these lines from Gemini to suppress it myself
%\begingroup % Start a local redefinition group for the LR bibliography
%\renewcommand\refname{} % Temporarily set the automatic title to empty
%\bibliographystyle{lrec2026-natbib}
%\bibliography{lrec2026-example,anthology}
%\endgroup % End the local redefinition group

\section*{Language Resource References}
\label{lr:ref}
%\bibliographystylelanguageresource{lrec2026-natbib}
%\bibliographylanguageresource{languageresource}

\end{document}